%% file: main.tex
\documentclass[conference,10pt,top=0.7in]{IEEEtran}
\IEEEoverridecommandlockouts

\usepackage{graphicx,color}%,spconf}
\usepackage{amsmath, amsthm, amsfonts, amssymb, amsbsy,nccmath}
\usepackage{algorithm}
\usepackage{enumerate}
\usepackage{lipsum}

\usepackage{textgreek}
\usepackage{textcomp}

\usepackage{algorithmic}
\usepackage[sort,compress]{cite}
\usepackage{epsfig}
\usepackage{epstopdf}
\usepackage{mathtools}
\usepackage{dsfont}
\usepackage{epstopdf}

\theoremstyle{definition}
\newtheorem{definition}{Definition}[section]

\usepackage[inline]{enumitem}   
\makeatletter
\newcommand{\inlineitem}[1][]{%
\ifnum\enit@type=\tw@
    {\descriptionlabel{#1}}
  \hspace{\labelsep}%
\else
  \ifnum\enit@type=\z@
       \refstepcounter{\@listctr}\fi
    \quad\@itemlabel\hspace{\labelsep}%
\fi}
\makeatother
\input{defines}

\makeatletter
\newcommand\fs@spaceruled{\def\@fs@cfont{\bfseries}\let\@fs@capt\floatc@ruled
  \def\@fs@pre{\vspace{0.5\baselineskip}\hrule height.8pt depth0pt \kern2pt}%
  \def\@fs@post{\kern1pt\hrule\relax}%
  \def\@fs@mid{\kern2pt\hrule\kern2pt}%
  \let\@fs@iftopcapt\iftrue}
\makeatother

\newcommand{\E}{\mbox{E}}

\newcommand{\bmzeta}{\boldsymbol{\zeta}}

\newcommand{\bit}{\begin{itemize}}
\newcommand{\eit}{\end{itemize}}

\newcommand{\mK}{\mathcal{K}}

\newcommand{\mG}{\mathcal{G}}

\newcommand{\bmz}{\boldsymbol{z}}

\newcommand{\bmg}{{\boldsymbol{g}}}

\newcommand{\bmX}{{\boldsymbol{X}}}

\DeclarePairedDelimiter\abs{\lvert}{\rvert}%

\newcommand{\bmZ}{{\boldsymbol Z}}

\usepackage{lipsum}
\restylefloat{algorithm}

\begin{document}
%\ninept
\linespread{0.9}

%\title{Convergence analysis of sparse Bayesian learning under approximate inference techniques}
%\name{Christo Kurisummoottil Thomas, Dirk Slock\thanks{\scriptsize EURECOM's research is partially supported by its industrial members:
%ORANGE, BMW,
%Sy\-man\-tec, SAP, Monaco Telecom, iABG,  and by the projects DUPLEX (French ANR) and MASS-START.}}
%\address{EURECOM, Sophia-Antipolis, France, Email:\{kurisumm,slock\}@eurecom.fr\\}
\title{ Neuro-Symbolic Artificial Intelligence (AI) for Intent based Semantic Communication\vspace{-4.5mm}}
\vspace{-8mm}
\author{\fontsize{1}{1}\selectfont
\IEEEauthorblockN{\fontsize{11}{12}\selectfont
Christo Kurisummoottil Thomas\IEEEauthorrefmark{1} and Walid Saad\IEEEauthorrefmark{2}
\vspace{0mm}}
\IEEEauthorblockN{\fontsize{10}{12}\selectfont
\IEEEauthorrefmark{1}\fontsize{9}{9}\selectfont Qualcomm, Finland and \\
\IEEEauthorrefmark{2}Wireless@VT, Bradley Department of Electrical and Computer Engineering, \\ Virginia Tech, Arlington, VA, USA, \\ \fontsize{8}{8}Emails: ckurisum@qti.qualcomm.com, walids@vt.edu\vspace{-6mm}
}}
\maketitle
\vspace{-12mm}
\begin{abstract}
Intent-based networks that integrate sophisticated machine reasoning technologies will be a cornerstone of future wireless 6G systems. Intent-based communication requires the network to consider the semantics (meanings) and
effectiveness (at end-user) of the data transmission. This is essential if 6G systems are to communicate reliably with fewer bits while simultaneously providing connectivity to heterogeneous users. In this paper, contrary to state of the art, which lacks explainability of data, the framework of \emph{neuro-symbolic artificial intelligence (NeSy AI)} is proposed as a pillar for learning causal structure behind the observed data. In particular, the emerging concept of generative flow networks (GFlowNet) is leveraged for the first time in a wireless system to learn the probabilistic structure which generates the data. Further, a novel optimization problem for learning the optimal encoding and decoding functions is rigorously formulated with the intent of achieving higher semantic reliability. Novel analytical formulations are developed to define key metrics for semantic message transmission, including semantic distortion, semantic similarity, and semantic reliability.  These semantic measure functions rely on the proposed definition of semantic content of the knowledge base and this information measure is reflective of the nodes' reasoning capabilities. Simulation results validate the ability to communicate efficiently (with less bits but same semantics) and significantly better compared to a conventional system which does not exploit the reasoning capabilities. 
\end{abstract}

\vspace{-1.5mm}
\section{Introduction}
\vspace{-1mm}

\indent  Future wireless systems (e.g., 6G) must be more judicious in what they transmit if they are to integrate time-critical autonomous system applications. Conventional wireless systems focus on reliably sending physical bits without emphasis on the \emph{semantic and effectiveness} layer, as pointed out by Shannon \cite{ShannonBSTJ1948}. Instead of transmitting the entire data, it is naturally more efficient in terms of delay, bandwidth utilization, and energy (without compromising the reliability), to only send information that is useful for the receiver \cite{StrinatiComNetworks2021}. This is central premise of so-called \emph{intent-based semantic communication (SC) systems} \cite{KountourisCommMag2021}. Intent-based networks are autonomous systems that define the behavior they expect from their network, e.g., “improving network quality,” for the system to then automatically translate it into real-time network action. 

Integrating the semantic and effectiveness aspects to create intent-based wireless networks requires a major paradigm shift \cite{StrinatiComNetworks2021,KountourisCommMag2021,Mehdi2021}. It particularly requires the transmit and receive nodes to move away from being just blind devices (that transfer data back and forth) towards brain-like devices capable to understand and reason over the data and how it gets generated. One promising approach here is combining knowledge representation and reasoning tools with machine learning. Once intelligence is embedded at transmitter and receiver, the communicating devices can sense (data acquisition), preprocess, and communicate efficiently, without unnecessary network bottlenecks (by sending large unnecessary data). Despite a surge in recent works here \cite{ChoiArxiv2022, LiuISIT2021,Mehdi2021, FarshbafanArxiv2022}, most of them fails to address some critical aspects in SC. In particular, these prior works typically ignore the reasoning behind the data generation. Instead, they limit the discussion to the computation of meaningful attributes which can describe some observed data. However, this is not very realistic to achieve the intent behind SC since the solutions of \cite{ChoiArxiv2022, LiuISIT2021,Mehdi2021,FarshbafanArxiv2022} are not generalizable enough to situations not encountered during training. 

\vspace{-0.5mm}
Moreover, the authors in \cite{ChoiArxiv2022} develop a unified
approach to study the interplay between semantic information, and communication through the lens of probabilistic logic (ProbLog). However, this solution requires adding an SC layer, which exchanges
logically meaningful clauses in knowledge bases (KBs) on top
of the existing technical communication (TC) layer. 
SC requires a fundamental paradigm shift with respect to (w.r.t) how the different components in the system are designed and, it cannot be relegated to being a separate layer aiding the technical layer. In \cite{Mehdi2021}, the authors show that infusing contextual reasoning for SC could reduce the effective number of bits transmitted without compromising reliability. Further, the work in \cite{MerluzziEuCNC2022} studies energy efficiency for SC while maintaining a desired delay and accuracy for edge inference. However, the artificial intelligence (AI) scheme of \cite{MerluzziEuCNC2022} lacks generalization capabilities which is vital for enhancing end nodes with reasoning capabilities. 

In contrast to this prior art, our main contribution is a rigorous framework for extracting the meaningful (at the receiver end) semantics hidden inside the data and for optimizing the transmitted message with high semantic reliability. Specifically, we consider an SC system in which the transmitter node sends the most meaningful message sequence that explains the observed data, to a receive node. For this system, to enable reasoning, we propose to use the emerging framework of \emph{neuro-symbolic AI} (NeSy AI) \cite{GarcezArxiv2020} and \cite{RaedtIJCAI2020}. NeSy AI combines two paradigms: neural and symbolic. The symbolic component characterizes the high level semantic representations while logical deduction and learning are achieved by a neural network component. A key benefit of NeSy AI to SC is an understanding of the representational value of
the symbolic manipulation of variables in logic and the compositionality of
language \cite{GarcezArxiv2020}.
In our SC system, the \emph{symbolic component} is represented by the KB available at the transmitting (speaker) and receiving (listener) nodes and constructed based upon the novel semantic language rules. The \emph{reasoning part} and the encoder/decoder functions are learned by a neural component called \emph{generative flow networks (GFlowNet)} \cite{BengioNeurIPS2021,DeleuArxiv2022}. For this system, our key contributions include:
\vspace{-1mm}\begin{itemize}
    \item Based on the rigorous real logic inspired concept of symbol groundings, we develop a novel semantic information measure based on the degree of confidence of the logical formulas represented by a KB at any edge node.
    \item We introduce the new concept of NeSy AI, to rationalize or add reasoning capabilities to end nodes in our SC  system. Then, we formulate an optimization problem whose goal is to choose the optimal transmit message  using a compact and concise objective function that is dependent on GFlowNet parameters. Our objective function is representative of two factors: 1) physical channel effects and imperfections in the KB at the listener and 2) causal structure learning of the Bayesian network, which generates the observed data.
    \item We solve the optimization problem of optimal encoder/decoder functions via training the GFlowNet. GFlowNet is augmented with the deep neural networks (DNN) representing the encoder and decoder functions. To our best knowledge, this is the first work in the literature that exploits NeSy AI to propose an intent-based SC model.  
    \item Simulation results show that enabling NeSy based SC helps in achieving reasonable reliability without losing the meaning or effectiveness at the listener, with fewer bits compared to conventional communication systems.
\end{itemize}

\vspace{-1.5mm}
\section{Proposed SC System Model}
\vspace{-1.5mm}

\begin{figure*}[ht]\vspace{-1mm}
 %\centerline{\includegraphics[width=8.3cm,height=5cm]{SSP_NMSE_GAMP_IF_SAVE}}
 %\centerline{\includegraphics[width=9cm]{NMSE_EM_LAMP_SBL}}
 \centerline{\includegraphics[width=6.5in,height=1.3in]{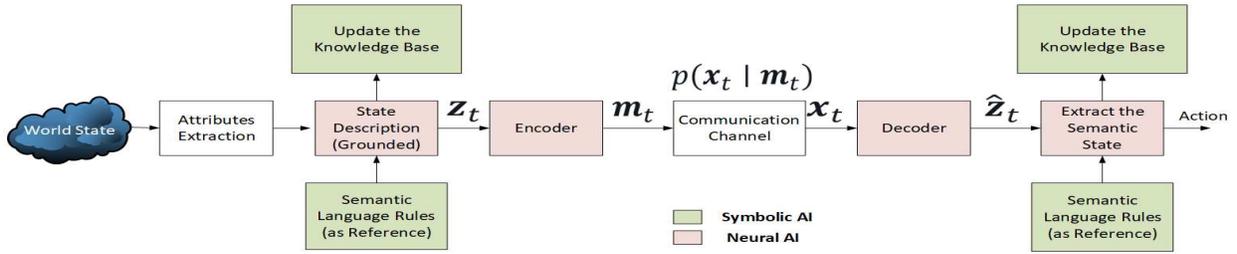}} \vspace{-3mm}
\caption{Overview of proposed SC model. Our focus is on 1) Learning of the state description, which is grounded on the real logic and formed based on the semantic language rules. 2) Encoder/decoder functions that translate the states to optimal physical messages with the objective of transmitting less but with semantic reliability close to 1. Objectives are achieved via the tools of NeSy AI.}
\label{Fig_SCModel}
\vspace{-7mm}
\end{figure*}
Consider a pair of speaker and
listener who wish to complete a set of sequential tasks, with a task denoted as $T$. Each
task is composed of multiple events $e\! \in\! \mathcal{A}$ (captures the state of the environment) observed sequentially by
the speaker. The speaker
must describe each observed event to a remote listener. In
response to the received information and according to the
events described by the speaker, the listener must take a specific
action to steer the system towards completing the ongoing
task. This is a central feature of intent-based networking. For example, consider a set of flying robots
 that share and distribute tasks wirelessly (e.g., networked
 robotics for autonomous road traffic control, which is the
 $T$ here). Each robot generates and sends updates of
 a continuous stochastic process (e.g., event $e$, a vehicle’s
 trajectory) to a remote tracking unit for real-time causal
 reconstruction. In this scenario, relatively static background information like the pedestrian crossing, objects in the surroundings, etc., can be assumed to be learned initially during the training phase (KB here). All the flying robots observing different regions of the environment may decide to transmit the entire acquired information (which may contain unnecessary background knowledge). Such an approach may create bottlenecks affecting the quality of real-time tracking intended to avoid any possible accidents. Instead, the speaker needs to transmit only the meaningful data. The speaker and the listener should then be able to deduce certain conclusions (logical formulas in our semantic language defined in Section~\ref{SemanticLanguage}) from the data and generalize to scenarios not encountered before during training. For example, a drunk person may be about to cross the road (not at a pedestrian crossing). By observing this person's behavior while off the road, the speaker and listener should be able to reason whether or not he may act abnormally and trigger a traffic hazard?. The same scenario may not have been part of the training period nor has it occurred before. If the future autonomous transportation systems have to be robust in all these scenarios, then the speaker and listener must be reasoning nodes. Moreover, when there are multiple robots who want to communicate, each of them has to be judicious in what gets transmitted, and, here, SC plays an inevitable role in achieving the intent behind the connected network of robots. 

 Acquiring, processing, and transmitting vast amounts of irrelevant (w.r.t the end-user) data will cause severe communication bottlenecks. A more efficient approach in terms of the transmission cost (proportional to the number of bits) would be to instill capabilities for the end nodes to deduce (not just reconstruct) logical conclusions from the data. This capability represents the \emph{reasoning} aspect of this work. Indeed, our envisioned SC model, shown in Figure~\ref{Fig_SCModel}, brings significant benefits to such a scenario. This involves several modules whose design choices are interdependent. Enhancing the end nodes with the KB (whose contents are constructed based upon semantic language) and reasoning helps to achieve the intent of transmitting less without reducing reliability. In the scope of this work, we primarily focus on causal structure learning (a so-called \emph{World Model} that captures the KB), and the optimal transmit message choice.

\vspace{-2mm}
\subsection{Semantic Language Construction}
\label{SemanticLanguage}
\vspace{-1mm}

For intent-based systems, it is desirable to transmit only the
meaningful representation of the event rather than the observed data. To represent the state
description of the event concisely, we must define a \emph{semantic language} whose rules are agreed upon between speaker and
listener. This language is used to represent the relevant
entities, attributes, and their relations. Considering the flying robots example, instead of sending all the acquired data, the systems can encode only the relevant objects (vehicles, people etc) and their relations in the environment and send them to the central controller,  saving bandwidth. We now ask: How can we map the events to the state description (which concisely represent the hidden meaning) using the rules in semantic language? Before answering this, we must introduce few preliminary concepts from Carnap's semantic language in \cite{Carnap52}, chosen as a basis here since it is the most rigorous framework that addresses the 1) semantic aspects of information through a logical language and 2) distinction between the amount of information in classical (Shannon's) and semantic sense. 

\textit{ Semantic Language Preliminaries:} We define a language $\mathcal{L}_n^{\pi}$, where $n$ denotes the number of \emph{entities} (things or objects or indivuduals of interest) and $\pi$ is the total number of \emph{attributes} (or properties of the entities) and \emph{relations} (between entities).  $\mathcal{L}$-formulas allow us to specify relational knowledge with variables, e.g., the atomic
formula $about\_to\_collide(Car, Bus)$ may state that the entity $Car$ may collide with $Bus$. This ends up as a warning to both vehicles in the network to take action such that high risk scenario is avoided. We define a KB as $\mathcal{K}\!=\!\{ \mathcal{E}, \mathcal{P}, \mathcal{R},  \mathcal{F}\}$, where $\mathcal{E}$, $\mathcal{P}$, $\mathcal{R}$, $\mathcal{F}$ are the sets of entities, predicates, relations and facts, respectively. A fact is a triplet $(h,r,t) \in \mathcal{F}$, where $h\in \mathcal{E}$ is the head of relation $r \in \mathcal{R}$ and $t\in \mathcal{E}$ is its tail. Note that we assume that the speaker shares the KB during the initial training phase (or is mutually learned). In every logical language, the facts or relations or predicates are represented through sentences, which are defined in the next few lines. In an atomic sentence $'Pr'$, an entity $r$ has the property $P$. For example, $P$ can correspond to a person's attributes e.g., weight, height etc. We also call an atomic sentence a term. Any sentence is either logically true, false or factual (indeterminate). As illustrated in \cite{Carnap52}, other molecular sentences are formed using the connectives such as $\{\lor,\land,\neg,\!\implies\!, \!\iff\!,\!\Leftrightarrow
\}$, that represent, respectively, disjunction, conjunction, negation, logical implication, bi-conditional (if and only if) and logical equivalence. We also define a state-description of the possible world as the conjunction of atomic sentences. While a state-description (that involves just entities and their attributes) may completely describes the possible discourse of a universe under question, it does not convey the entire semantic information associated with the event $e$. For an accurate description of the semantics, we also look at the relations between different entities under question in Section~\ref{GFlowNet}, an aspect that was not studied in Carnap's work \cite{Carnap52}. 

\vspace{-3mm}
\subsection{GFlowNet for Relational Structure Learning}
\label{GFlowNet}
\vspace{-1mm}

One approach for obtaining the relations between various entities which are part of the state description is to learn the \emph{causal structure} which leads to the data generation. The most popular approach here is to investigate the posterior distribution of the entities given the observed event. This probabilistic structure can be either assumed to follow a particular approximate distribution as in variational inference techniques or the exact (or close to optimal) distribution can be learned using an AI method which is generalizable to out of order distributions. This is in contrast to the prior art in SC (e.g., \cite{Mehdi2021} and \cite{ChoiArxiv2022}),  where the AI methods lack generalizability.
 The recently proposed GFlowNet \cite{BengioNeurIPS2021} are one such class of generative probabilistic models over a discrete and structured sample space $\mathcal{S}$. They have the  capacity to generalize
to states not encountered during training. Assume that the attributes ($s_i \in \mathcal{S}$) corresponding to an event are extracted, we look at how to explain this particular scene to a listener. To do so, we define the event as a sequence of states (represented as nodes in a \emph{directed acyclic graph (DAG)} $G$), with a possible edge between two nodes if there is a relation between these two entities. Starting with an initial state $s_0$, we seek to estimate the following posterior distributions:
 \vspace{-2mm}\beq  \vspace{-2.5mm}
 \begin{array}{l}\vspace{-1mm}
 p(s_0,\cdots,s_f\mid e) = \prod\limits_{i} p(s_i \mid \textrm{pa}(s_i)), 
 \end{array} \vspace{-2mm}
 \eeq
where $\textrm{pa}(s_i)$ represents all the parent nodes of the node $s_i$. These states $s_i$ which represent accurately an event are the entities and attributes in our semantic language. Our aim here is to learn the causal structure leading to every event. We represent the state description $\bmg= \left[s_0\cdots s_f\right]^T\!$ as the sequence of states which describe an event. Moreover, we also define an $N\!\times\!N$ adjacency matrix $\bmW$, where $N$ is the optimum set of entities and attributes which can describe a particular event. Each entry $\bmW_{i,j}$ of $\bmW$ is a binary value which indicates whether a directed edge is present from node $s_i$ to $s_j$. We define $\btheta$ as the set of GFlowNet parameters, to be optimized.

%\begin{figure}[h]
 %\centerline{\includegraphics[width=8.3cm,height=5cm]{SSP_NMSE_GAMP_IF_SAVE}}
 %\centerline{\includegraphics[width=9cm]{NMSE_EM_LAMP_SBL}}
 %\centerline{\includegraphics[width=3.6in,height=2in]{GC22_GFlowNet_StateFlow}}
%\caption{State Transition Diagram in GFlowNet}
%\label{GC22_GFlowNet_StateFlow}
%\vspace{-2mm}
%\end{figure}
The different GFlowNet components are provided in \cite[Figure 2]{DeleuArxiv2022}. Each directed edge (between $s_i$'s) is embedded using its source and target's embeddings (\emph{grounded} value), with an additional vector indicating whether the edge is present in $G$. These embeddings are fed into a linear
transformer, with two separate output heads. The first head (above) gives the probability of adding a new edge $P_{\btheta}(G^{\prime}
\!\mid \!G, \neg s_f )$, using the mask $M$ associated with $G$ to filter out invalid actions. $G^{\prime}$ is the newly formed DAG. The second head (below) gives the probability of terminating the trajectory, $P(s\!=\!s_f\!\mid\! G)$. Hence, we now created a probabilistic graphical model to represent the causal structure explaining the data using the novel explainable AI tools of GFlowNet.  %In Section~\ref{Encoder}, we look at how we optimize the parameters of the GFlowNet, such that we are able to transmit the semantic content reliably to the listener with an accurate description of the system state.  

\vspace{-1.5mm}
\subsection{Grounding of Symbolic Elements}
\label{Grounding}
\vspace{-1.5mm}

In Section~\ref{GFlowNet}, we looked at how to understand the relations among the entities. Next, in order to physically transmit the causal sequence $\bmg$, we need to construct a sentence using the semantic language rules defined. For example, referring to our flying robots, the causal sequence learned by any robot may be ``car, bus lane" related by the attribute ``heading". What we intend to transmit would be the sentence ``The car heading in the bus lane". Our aim here is to perform \emph{symbol grounding} which is a symbolic AI process that can translate the logical symbols and the relations learned into meaningful representations in real logic. The grounded representations get communicated to the listener. Mathematically, this represents a transition from the causal structure to the grounded state description, $\mathcal{G}\!\!:\bmg \rightarrow \bmz$. One of the disadvantage with the existing inductive and deductive logic based representations considered in \cite{ChoiArxiv2022} and \cite{Carnap52} is that such sub-symbolic models lack comprehensibility. From an SC point of view, all the above logical representations and information measures in \cite{Carnap52} do not have any physical relevance unless we have a \emph{grounded representation of the symbols} presented. In contrast, symbolic AI is built upon rich, high-level representations of the possible world that use human-readable symbols. In our KB, we utilize a fully differentiable
logical language, called \emph{real logic}. The elements of this first-order logic signature
are grounded onto data using neural computational graphs and first-order fuzzy logic
semantics, motivated by emerging AI techniques such as \cite{Badreddine2022}. Since we are interested in learning and reasoning in real-world scenarios where
degrees of truth are often fuzzy and exceptions are present, formulas can be partially true, and therefore we adopt \emph{fuzzy
semantics}. Mathematically, this means that we represent the degree of confidence in a truth or relation by a real number $\in [0,1]$.

An optimal grounding can also \emph{minimally} (in minimum possible dimensions) represent what needs to be physically transmitted given that listener also has the same KB. Corresponding to the KB defined $\mK$, we define a \emph{grounded KB} $\mG$, representing the logical symbols in $\mK$ grounded to real logic. Every entity, term, relational symbol or predicates, and the state description will be interpreted as a tensor of real values. The benefit of a tensor-based representation for the symbols is that it can belong to any type, or a heterogeneous combination. 
We define a key mapping, $\mathcal{D}: \mathcal{F} \rightarrow \mathcal{D}$, from the domain of logical symbols to the real logic. $\mathcal{D}$ is the domain of either entity or attribute or relation or facts. For example, $\mathcal{F}$, may be \textit{people} or \textit{vehicles} or relation \textit{about\_to\_collide} based on our sample KB. A grounding for a logical language $\mathcal{L}_n^{\pi}$ on $\mathcal{D}$ provides the interpretation of both the domain symbols
in $\mathcal{D}$ and the non-logical symbols in $\mathcal{L}_n^{\pi}$.
\vspace{-0.5mm}\begin{definition}\vspace{-1mm}
A \emph{grounding} $\mathcal{G}$ associates to each domain $D \in \mathcal{\bmD}$ a set $\mathcal{G}(D) \in \bigcup_{n_1 \cdots n_d \in N^*} \mathcal{R}^{n_1\cdots n_d}$. Without loss of any generality, we can also assume that the symbols belonging to the same domain have their grounding represented by a tensor of the same dimensions. For concatenation of different entities, their attributes and predicate symbols, where $\small D_i \in \mathcal{\bmD}^*$, $\mathcal{G}(D_1,\cdots,D_n) \!=\! \bigtimes_{i=1}^n \mathcal{G}(D_i) \!=\!\mathcal{G}(D_1) \bigtimes \cdots \bigtimes \mathcal{G}(D_n) $. Here, $\mathcal{\bmD}^*$ is the Kleene star of $\mathcal{\bmD}$.\vspace{-2mm}
\end{definition}\vspace{-0.5mm}
We further elucidate what grounding implies in the case of various symbols. A grounding assigns to each constant or entity $c$, a tensor $\mathcal{G}(c)$ in the domain $\mathcal{G}(D(c))$. %It assigns to a variable $x$ a finite
%sequence of tensors $d_1 \cdots d_k,$ each in $\mathcal{G}(D(x))$. These tensors represent the instances of $x$.  
A grounding assigns to a function symbol $f$ a function taking tensors
from $\mathcal{G}(D_{\textrm{in}}(f ))$ as input, and producing a tensor in $\mathcal{G}(D_{\textrm{out}}(f ))$ as output. Finally, a grounding assigns to a predicate symbol
$p$ a function taking tensors from $\mathcal{G}(D_{\textrm{in}}(p))$ as input, and producing a truth-value in the interval $[0, 1]$ as output. Further, if a grounding depends on a set of DNN parameters $\btheta$, then we represent it as $\mathcal{G}_{\btheta}(\cdot)$.

\vspace{-2.5mm}
\subsection{Semantic Information Content of the KB}
\vspace{-1.5mm}

Next, we quantify the semantic information conveyed by a KB, which, in real logic, represents the average degree of confidence across all the logical formulas included in the same. First, we define the equivalent notion of the KB in our semantic language in the real logic as follows:
\vspace{-0.5mm}\begin{definition}\vspace{-1.5mm}
\emph{The theory of real logic} is a triple \emph{$\mathcal{T} = (\mathcal{K},\mathcal{G}_{\theta}(\cdot),\bTheta)$}, where $\mathcal{K}$ is the KB which represents the logical symbols $S = \mathcal{E}\cup\mathcal{P}\cup\mathcal{R}\cup\mathcal{F}$, $\mathcal{G}_{\theta}(\cdot)$ represents the grounding on the symbols and $\theta_s \in \bTheta_s \subset \bTheta$ represents the parameters (learned through DNN) for the grounding. \vspace{-2mm}
\end{definition}
Before proceeding further with semantic information content definition, we define the grounding for the truth value or degree of confidence in a particular logical formula. %Here, we may use quantifiers such as $\{\forall, \exists\}$. %One such example of a logical formula  would be the statement $\mbox{``All the students are football lovers."}$. 
Different implementations, e.g., mean, minimum and harmonic mean, can be used to implement the universal quantifiers' aggregation. However, it may be quite difficult to find a universal aggregator operator (see \cite{Badreddine2022}) which conveys accurate semantic information in all cases. E.g., %considering an image interpretation case, one possible logical formula could be $\phi(x) = $``All cats have a tail."
considering our robotics example, one possible logical formula can be $\phi(x) = $``Majority of cars in the road are electric".
Here, an appropriate aggregate can be the mean function. The more positive examples you have, the higher the truth value of $\phi(x)$. %However,  if $\phi(x) =$``All people are football lovers", then one contradictory example is enough to prove that $\phi(x)$ is logically-false. Hence, an appropriate aggregator here would be $\mathcal{A}(\forall x \phi(x)) = \min_{x\in People} \phi(x)$. 
Further, we utilize the aggregator concept to state the semantic content of a theory.
\begin{definition}
\vspace{-1.5mm}
 The \emph{semantic content of a theory $\mathcal{T}$} can be represented using the aggregation operator $\mathcal{A}$ and it is defined as the following semantic measure function (sum of the grounding for the aggregation across all logical formulas included in $\mathcal{K}$)
\vspace{-1mm}\beq\vspace{-0mm}
%\begin{align}
S(\mathcal{T}) =  \textstyle\sum\limits_{\forall \phi \in \mathcal{K} }-\log \left( \mathcal{A}(\phi)\right).
%\end{align}
\vspace{-2mm}
\label{eq_semanticInf}
\eeq
\vspace{-3mm}
\end{definition}
\vspace{-2.5mm}
This measure function is motivated from the information measures defined in \cite{Carnap52}. The simplest semantic quantification in \eqref{eq_semanticInf} goes beyond classical Shannon's TC-focused theory by assuming that both the transmitter and receiver are intelligent agents (brought by the triple $\mathcal{T}$).

\vspace{-1mm}
\section{Encoder/Decoder Problem 
Formulation: Optimizing Transmit Message}
\label{Encoder}
\vspace{-1mm}

In Section~\ref{Grounding}, we looked at grounded representation of the state description, denoted by $\bmz$. However, one question we must address next: What is an optimal transmit message scheme that enables the listener to reliably reconstruct the semantic aspect considering that it also owns a KB that may be distorted, compared to speaker's KB? For example, in the robotics case, the listener should be able to question himself on whether the decoded message corresponds to a situation that can cause an accident to the car, and hence a warning needs to be sent.  Our challenge here is to encode the grounded state description $\bmz$ onto the transmit signal vector $\bmm$ given the channel realization $p(\bmx\mid \bmm)$, where $\bmm \in \mathcal{M} (\subset R^K)$ is the encoded message to be transmitted. Given $\bmz$ which accurately describes the system state, our key goal is to optimize the message while leveraging reasoning capabilities at speaker and listener. The speaker's encoder is a DNN with multiple
dense layers followed by a normalization layer that meets the physical constraints on $\bmm$, \cite{TOSheaTCN2017}. We write
$\bmm \!=\! f_{\bmeta}(\bmz)$,
where $\bmeta$ represents the DNN parameters.
We emphasize that, during the entire duration of the particular task $T$ at hand, the DNN weights $\bmeta$ are considered constant. At the listener, we represent the decoder using a similar DNN with parameters $\bmzeta$ and $\widehat{\bmm} \!= \!g_{\bmzeta}(\bmx)$.
Here, $\bmm, \widehat{\bmm} \in \mathcal{R}^K$. Hereinafter, we assume that the state description $\bmz$ is given (causal structure using GFlowNet) for the particular event observed. We then design an optimal choice for the message in our SC system. 

The communication of the current task can be represented using a sequence of messages, ${\bmm_1,\!\cdots\!,\bmm_L}$. Upon extraction of the semantic state $\bmzh_t$ at channel use $t,$ the listener takes a $d$-dimensional action vector $\bma_t \!\in\! \mathcal{R}^d$. This may physically represent the action taken by the listener, which gets reflected in the next sequence of events observed by the speaker.  Assume that $\mathcal{T}_t$ is the KB (evolving during data transmission) at the speaker. If the transmit message is communicated error-free, then, the listener acts based on policy $\small p(\bma_t\! \mid \!\bmm_t, \mathcal{T}_{t-1})$. Given the same set of semantic messages, the decoder function which applies the transformation $\bmm_t \rightarrow \bmz_t\!\mid\! \mathcal{T}_{t-1}$, where $\bmz_t$ is the state description of the environment is error-free. When the channel is erroneous as captured by the probability distribution $p(\bmx_t\!\mid\! \bmm_t)$, the transformation $\bmx_t\! \rightarrow \!\widehat{\bmz}_t\!\mid\! \widehat{\mathcal{T}}_{t\!-\!1}$ will be imperfect, where $\widehat{\mathcal{T}}_{t-1}$ is the evolving KB at the listener. Here, $\bmx_t$ is the listener's received signal, after undergoing the channel effect via $p(\bmx_t\!\!\mid\! \!\bmm_t)$. Next, we define key metrics needed to formulate our semantic message transmission problem. Let $S_t(\mathcal{{T}})$ and $S_t(\mathcal{\widehat{T}})$ be the amount of semantic information conveyed by the speaker's and listener's KB, respectively. At any instance $t$, the amount of semantic information is:
\vspace{-1.5mm} 
\beq
\vspace{-3mm} 
\begin{array}{l}
\small
S_t(\mathcal{T}) = S_{t-1}(\mathcal{T}) + \sum\limits_{\forall \phi \in  (\mathcal{K}_{t-1} \cup \Delta_{\mathcal{K}})}\mathcal{A}\left(\mathcal{G}_{\theta}(\phi)\right),
\end{array}
\eeq
where $\Delta_{\mathcal{K}_t}$ represents the changes in the semantic KB after the state gets updated as $\bmz_t$. %Note that the observed event $e_t$ may convey additional semantic information (represented by a set of $\phi_o \in \mathcal{K}_{t-1}$). However, since those are already part of the KB at the speaker (at the listener if he has a perfect version of the same), this set should not be part of the transmitted state description. In a classical communication system, this need not be the case since whatever data is acquired will be transmitted (but stale from the listener's point of view).

\vspace{-2mm}
\subsection{Semantic Distortion and Semantic Similarity}
\label{SD}
\vspace{-1mm}

The channel imperfections and the inaccuracy of the KB at the receiver results in a \emph{semantic distortion}, which can be captured by the squared error between speaker's transmitted message ($\bmm_t$) and the listener's learned message ($\widehat{\bmm}_t$):
\vspace{-1mm}
\beq
\vspace{-1mm}
%\begin{align}
E_t(\bmm_t,{\widehat{\bmm}}_t) = \abs{\bmm_t-{\widehat{\bmm}}_t}^2.
%\end{align}
\vspace{-1mm}
\eeq
We also define the error measure in terms of the difference in semantic information conveyed by the speaker and that learned by the listener,  $E_t(\mathcal{T}_t,\mathcal{\widehat{T}}_t) = \abs{S_t(\mathcal{T})-S_t(\mathcal{\widehat{T}})}^2.$
The semantic distortion is helpful to quantify how much semantic information the listener can extract from an inaccurate version of the message received.

Next, we define the concept of \emph{semantic similarity} which is not the bit error rate (BER) of classical systems and is required for defining the semantic reliability constraint. In a sense, we may send a message that experiences a high BER, but the semantics (represented by the semantic similarity) may be sufficient to recover the intended meaning. For example, say the listener receives the English sentence ``Aln the vhicles have sped lss thn 50 km/hr" instead of ``All the vehicles have speed less than 50 km/hr" (actually intended by the speaker). The received text has high BER (called an error at syntactic level \cite{StrinatiComNetworks2021}), i.e., typos, but its semantic content is rich and can be recovered. Several sentences can correspond to the same state description in our semantic KB model. The decoded message at the listener restricted to this semantic message space (say represented as $\mathcal{M}_t$ corresponding to message $\bmm_t$) can represent the same semantic content and they are semantically similar. Hence, we can achieve the same reliability. This semantic message space (representing semantic similarity) corresponding to $\widehat{\bmm}_t$ can be described as using the following measure space:
\vspace{-2mm}
\beq
\vspace{-1mm}
\begin{array}{l}
E_t(\bmm_t,{\widehat{\bmm}}_t)  \leq \delta,\,\, \mbox{s.t.}\,\, E_t(\mathcal{T}_t,\mathcal{\widehat{T}}_t) = 0.
\end{array}
\label{eq_semantic_meas_space}
%\vspace{-0mm}
\eeq
Compared to the classical BER measure, the semantic distortion $E_t(\bmm_t,{\widehat{\bmm}}_t)$ need not be restricted by an arbitrarily low value to achieve a reliability very close to $1$. Further, we introduce metric $C_t$, interpreted as the causal influence on SC, and it captures the causal impact of the speaker's message as observed through a channel with a response characterized using $p(\bmx_t\mid \bmm_t)$. This also quantifies the discrepancies between the KB of the speaker and that of the listener. We define $\small  C_t(\bmm_t,\mathcal{T}_t,\mathcal{\widehat{T}}_t)\! \!= \!\!D_{KL} \left(p(\bma_t \!\!\mid\!\! \bmm_t, \mathcal{T}_{t-1}) \,||\, \sum\limits_{\bmx_t} p(\bmx_t\! \mid\! \bmm_t )p(\bma_t\! \!\mid\! \!\bmx_t, \mathcal{\widehat{T}}_{t-1})\right) $, where $D_{KL} (p||q)$ represents the KL divergence between $p$ and $q$.
%\vspace{-1mm} \beq\vspace{-3mm} \small
%\begin{array}{l}
%C_t(\bmm_t,\mathcal{T}_t,\mathcal{\widehat{T}}_t) = \\ D_{KL} \bigleft(p(\bma_t \mid \bmm_t, \mathcal{T}_{t-1}) \,||\, \sum\limits_{\bmx_t} p(\bmx_t \mid \bmm_t )p(\bma_t \mid \bmx_t, \mathcal{\widehat{T}}_{t-1})\bigright)  \nonumber
%\end{array}\vspace{-1mm}
%\eeq

\vspace{-1mm}
\subsection{SC Message Transmission Problem Formulation}
\vspace{-1.5mm}

Utilizing the semantic metrics defined above, we can now rigorously formulate our objective. The speaker aims to choose an optimal transmit message to represent the semantic information at the transmit side to achieve a desired semantic reliability. Using the measures discussed in Section~\ref{SD}, \emph{semantic reliability} (i.e.,$=p\left(E_t({\bmm}_t(\bmeta),{\widehat{\bmm}}_t(\bmzeta) ) < \delta \right)$) is defined so that the listener can reliably reconstruct all the logical formulas contained in the decoded state description. Compared to reliability measures in classical communication theory, in semantics, we can recover the actual meaning of the transmitted messages even with a larger BER, as long as the semantic distortion is within a limit. We can now formally pose our problem, as follows:
\vspace{-1mm}\begin{subequations}\vspace{-1mm}
\begin{align}
 \min \limits_{\btheta, \bmeta,\bmzeta}
\bmC_t(\bmm_t(\bmeta),\mathcal{T}_t(\btheta),\mathcal{\widehat{T}}_t(\btheta)) + L_{\btheta}(\tau) \tag{6}  \\
\mbox{s.t}\,\,\, p\left(E_t({\bmm}_t(\bmeta),{\widehat{\bmm}}_t(\bmzeta) ) < \delta \right) \geq 1-\epsilon,\vspace{-1mm}
%\end{array}
\end{align}
\label{GFlowNetLossFunc}
\vspace{-1mm}
\end{subequations}\!\!\!where $\delta, \epsilon$ can be arbitrarily small values greater than 0. The loss function to optimize the GFlowNet parameters for optimal grounding (to represent the logical formulas) is expressed as
\vspace{-1.5mm}\beq
\vspace{-1.5mm}\small
\begin{array}{l}
\!L_{\btheta}(\tau) = \!\! \sum\limits_{s^{\prime} \in \tau \neq s_0}\!\!\left(\!\sum\limits_{s,a:\Pi(s,a)=s^{\prime}}\!\!\!\!F_{\btheta}(s,a)-R(s^{\prime})-\!\!\!\sum\limits_{a^{\prime}\in \mathcal{A}(s^{\prime})}F_{\btheta}(s^{\prime},a^{\prime})\!\right)^2
\end{array}
\eeq
Here, $\Pi(s,a)$ represents the state transition in the DAG. $R(s^{\prime})$ can be selected as a metric which is inversely proportional to the average number of bits to represent any grounding (for a logical formula) which passes through the state $s^{\prime}$. The flow $F_{\btheta}(s,a)$ is parameterized by the GFlowNet neural network and is computed during the training. 
\setlength{\textfloatsep}{0pt}
\begin{algorithm}[t]
\caption{NeSy Approach towards Semantic Transmission - Training Phase (Evolving KB) }\label{alg_1}%\vspace{-1mm}
\fontsize{7.5}{9}\selectfont \textbf{Given:} $\mbox{The KBs at the speaker and listener, at training instance}\\ \mbox{   }t-1,\mathcal{T}_{t-1},\mathcal{\widehat{T}}_{t-1}, \,\mbox{channel transition probability} \,\,p(\bmx_t\mid\bmm_t)$.\\
\textbf{Initialize:} \mbox{Adjacency matrix } $\bmW^{(0)}$ \,\,\mbox{as all zeros matrix}. Mask $\bmM^{(0)}$ to be constrained to have no loops. \;\\
\textbf{Set:} $\mu = 0.4$.
\begin{algorithmic} \vspace{-1mm}
\STATE \hspace{0.05cm}\textbf{ for minibatch b=1:B}
\STATE \hspace{0.4cm} Sample minibatch ($b$) of $N$ events from the training set, $e \sim D$. \\
\STATE \hspace{0.4cm} \textbf{for each event in $b$ repeat until terminal state is reached}\\
\STATE \hspace{0.7cm} Compute the  $P(G^{(t)}\mid G^{(t-1)},\neg s_f)$ and  $P(s_f\mid G^{(t-1)})$\\
\STATE \hspace{0.7cm} if ($P(s_f\mid G^{(t-1)}) > \mu$) \\
\STATE \hspace{1.1cm} exit the training for the event
\STATE \hspace{0.7cm} else
\STATE \hspace{1.1cm} Update the GFlowNet parameters, $\bmM^{(t)}$, and $\bmW^{(t)}$.
\STATE \hspace{1cm} Update the NN parameters for the encoder and decoder.
\STATE \hspace{0.4cm} \textbf{end for}
\STATE \hspace{0.05cm} \textbf{end for}\\
Use the learned GFlowNet and encoder/decoder weights for the data transmission.
\end{algorithmic}
\label{algo1}  \vspace{-1.5mm}
\end{algorithm}
\vspace{-1mm}
\subsection{Proposed NeSy AI Algorithm for Transmit Message Design }
\vspace{-1mm}

Our objective function \eqref{GFlowNetLossFunc} is a novel rigorous formulation that captures accurately the semantic reasoning capabilities associated with the end nodes. To solve \eqref{GFlowNetLossFunc} whose intent is to achieve high semantic reliability, we use the backpropagation to fine tune the GFlowNet and encoder/decoder DNN weights. Our objective function in \eqref{GFlowNetLossFunc} is convex. This can be easily proven since KL divergence is convex and the convergence properties of GFlowNet are already proved in \cite{BengioNeurIPS2021}. Hence, the convergence to any global minimum solution is guaranteed for our solution. We train over several mini-batches iteratively and hence the optimization of the network parameters are done over an $\small \E_{e\sim D}[\bmC_t(\bmm_t(\bmeta),\mathcal{T}_t(\btheta),\mathcal{\widehat{T}}_t(\btheta)) + L_{\btheta}(\tau)]$. Moreover, we consider our KB to be evolving which means that during the every training instance, terminal state probability $p(s_f\mid G)$ gets updated. The algorithmic details of the proposed NeSy AI based SC system is described in Algorithm~\ref{alg_1}. 

%\vspace{1mm}
%\setlength{\textfloatsep}{0.1cm}% Remove \textfloatsep
%\setlength{\floatsep}{0.1cm}
 \vspace{-1mm}
\section{Simulation Results}
 \vspace{-1mm}

We followed the experimental setup of \cite{DeleuArxiv2022} for the data generation. In each
experiment, a random graph $G$ was generated from the Erd\"os-R\'enyi (ER) random graph model. Given $G$, we uniformly assigned random edge weights to obtain a weight matrix $\bmW$. Given $\bmW$, we sampled $\bmX = \bmW^T \bmX + \bmZ$, where $\bmZ$ is a Gaussian noise. This generated data is fed to the GFlowNet, which learns the causal structure forming the data $\bmX$. The training phase is performed using 25,000 different observations, and 100 minibatches are considered. The GFlowNet and encoder/decoder are weights that are learned using this training phase. In the actual transmission phase, this same model is considered. From an SC perspective, these measurements $\bmX$ represent the data received from $N$ ($=5$ here) different sensors. At the listener, based on the quantized levels ($4$ possible levels considered) of the measurement vector decoded, different actions are taken. For simplicity, we skip the attribute extracton step here. The learned graphical structure is represented by the adjacency matrix $\bmW$, whose entries are binary. Further, the compressed representation for the optimal message $\bmm_t$ is computed and sent across a binary symmetric channel (BSC), with crossover probability as $p$. 

\begin{figure}[t]
%\vspace{-1mm}
 %\centerline{\includegraphics[width=8.3cm,height=5cm]{SSP_NMSE_GAMP_IF_SAVE}}
 %\centerline{\includegraphics[width=9cm]{NMSE_EM_LAMP_SBL}}
 \centerline{\includegraphics[width=3.3in,height=1.5in]{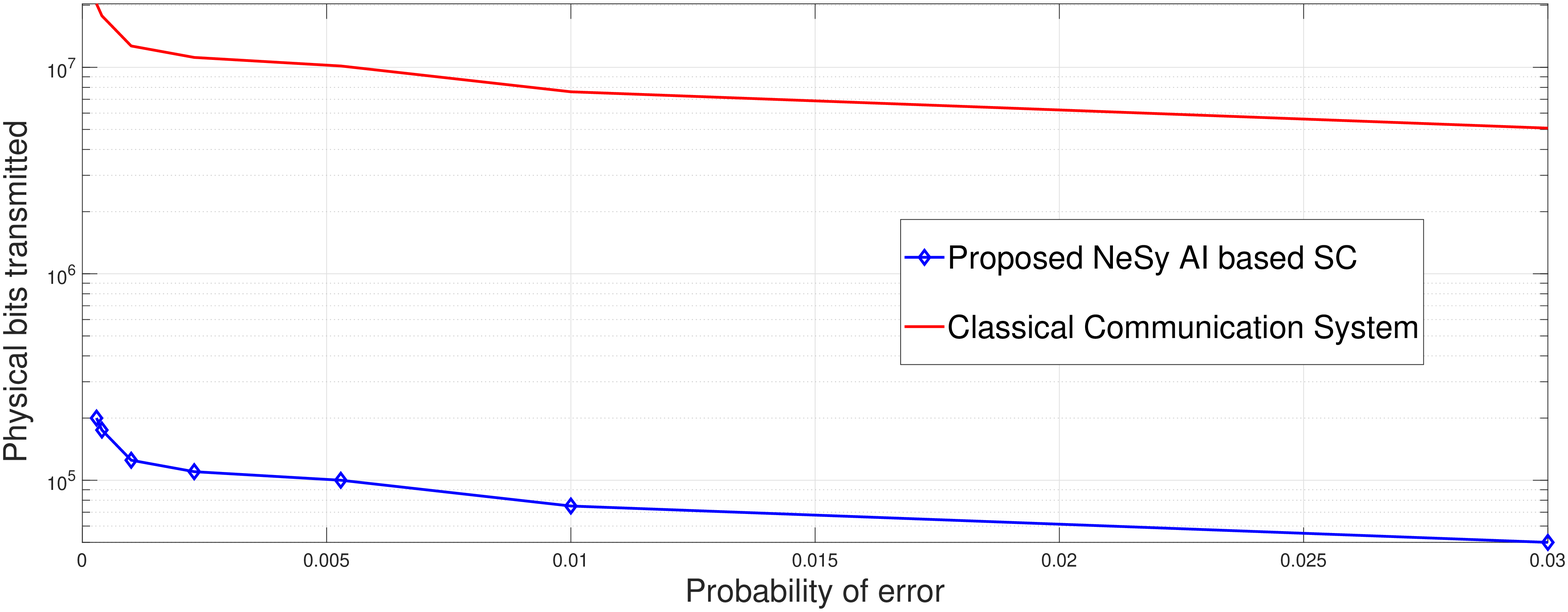}}\vspace{-2.5mm}
\caption{NeSy AI based Semantics vs Classical Communication.}
\label{Semantics}\vspace{-3mm}
%\vspace{-6mm}
\end{figure}

Figure~\ref{Semantics} shows that an SC system based on NeSy AI can be more efficient (in terms of amount of bits) to convey the same amount of semantic information compared to the conventional communication system. In Figure~\ref{Semantics}, we analyze the number of bits transmitted to get a particular probability of error (for 10000 events).  The amount of bits transmitted are reduced by a factor of 100 for the semantic system compared to classical system (without reasoning part) which explains the significance of our proposed methodology.

\begin{figure}[t]\vspace{-2mm}
 %\centerline{\includegraphics[width=8.3cm,height=5cm]{SSP_NMSE_GAMP_IF_SAVE}}
 %\centerline{\includegraphics[width=9cm]{NMSE_EM_LAMP_SBL}}
 \centerline{\includegraphics[width=3.3in,height=1.5in]{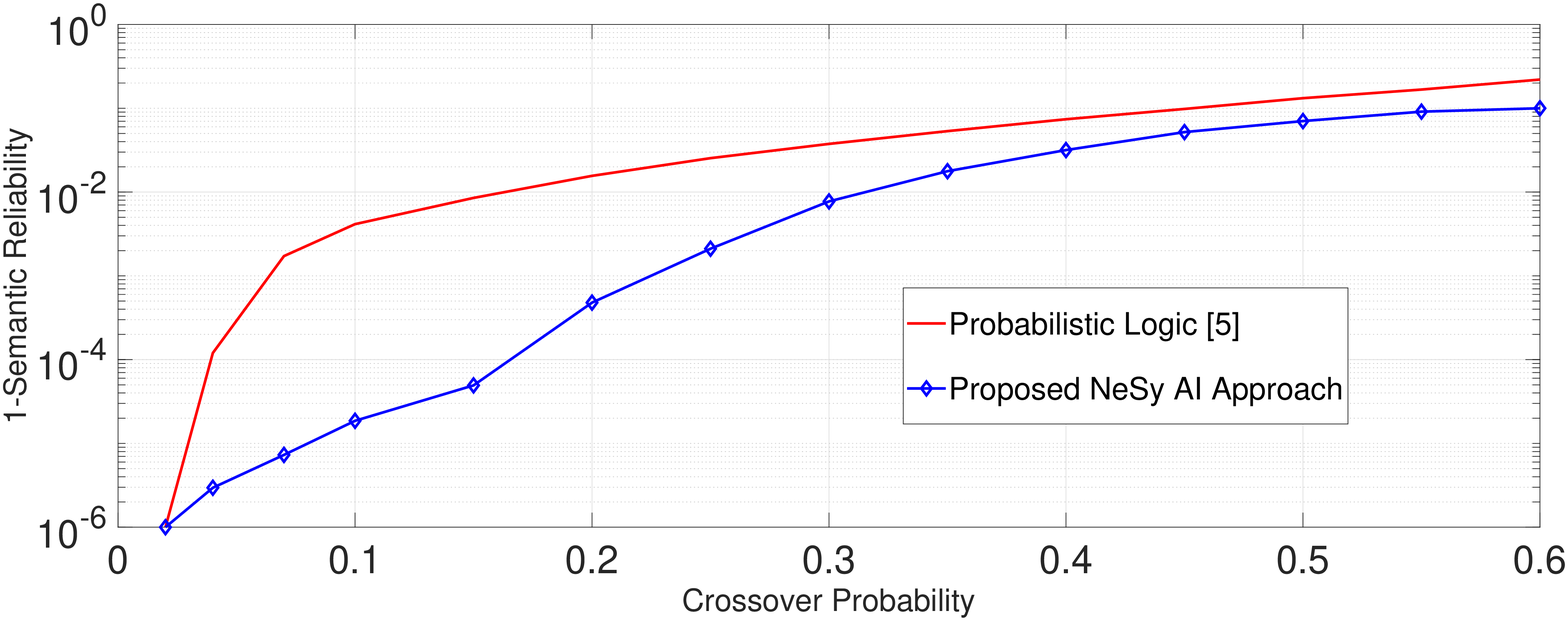}}\vspace{-3mm}
\caption{NeSy AI based Semantics vs Probabilistic Logic.}
\label{NeSyvsPbLog}\vspace{-1mm}
%\vspace{-6mm}
\end{figure}

In Figure~\ref{NeSyvsPbLog}, the performance of NeSy AI based SC system shows significant improvements compared to state of the art ProbLog based SC system. The evaluation is performed based on the decoding error probability (or $1-$ semantic reliability) vs $p$ for the BSC.


 \vspace{-2mm}
\section{Conclusion}
 \vspace{-2mm}

In this paper, we have introduced a novel SC model called NeSy AI to bring intelligence to the end nodes. In this system, the symbolic part is elucidated by a KB and the reasoning enabled by the GFlowNet. We have formulated an optimization problem for causal structure learning from the data and optimal encoder or decoder functions. Our simulations show significant gains in the amount of physical bits transmitted to convey the same meaning reliably compared to a classical network.
 \vspace{-1mm}
\bibliographystyle{IEEEbib}
\vspace{-1mm}
\bibliography{main}
%\vspace{-1mm}
\end{document}

%% file: defines.tex
\newcommand{\beq}{\begin{equation}}
\newcommand{\eeq}{\end{equation}}

\def\adots{\mathinner{\mskip0mu\raise0pt\vbox{\kern7pt\hbox{.}}\mskip3mu
          \raise4pt\hbox{.}\mskip3mu\raise8pt\hbox{.}\mskip0mu}}

\newcommand{\bmW}{{\boldsymbol W}}

\usepackage{bm}

\newcommand{\bmx}{{\bm x}}

\newcommand{\bmzh}{\widehat{\bmz}}

\newcommand{\bmM}{{\bm M}}

\newcommand{\bmC}{{\bf C}}

\newcommand{\bmD}{{\boldsymbol {D}}}
\newcommand{\bmm}{{\boldsymbol {m}}}

\newcommand{\bma}{{\bm a}}

\newcommand{\bTheta}{\boldsymbol{\Theta}}
\newcommand{\btheta}{\boldsymbol{\theta}}

\newcommand{\bmeta}{\boldsymbol{\eta}}

%% file: main.bbl
\begin{thebibliography}{10}

\bibitem{ShannonBSTJ1948}
C.~E. Shannon,
\newblock ``{A Mathematical Theory of Communication},''
\newblock {\em in The Bell System Technical Journal}, vol. 27, no. 3, pp.
  379--423, Jul. 1948.

\bibitem{StrinatiComNetworks2021}
E.~C. Strinati and S.~Barbarossa,
\newblock ``{6G Networks: Beyond Shannon Towards Semantic and Goal-Oriented
  Communications},''
\newblock {\em Computer Networks, Elsevier}, vol. 190, 2021.

\bibitem{KountourisCommMag2021}
M.~Kountouris and N.~Pappas,
\newblock ``Semantics-empowered communication for networked intelligent
  systems,''
\newblock {\em IEEE Communications Magazine}, vol. 59, no. 6, pp. 96--102, 202.

\bibitem{Mehdi2021}
H.~Seo, J.~Park, M.~Bennis, and M.~Debbah,
\newblock ``{Semantics-Native Communication with Contextual Reasoning},''
\newblock {\em arXiv preprint arXiv:2108.05681}, 2021.

\bibitem{ChoiArxiv2022}
J.~Choi, S.~W. Loke, and J.~Park,
\newblock ``{A Unified View on Semantic Information and Communication: A
  Probabilistic Logic Approach},''
\newblock in {\em In: arXiv preprint arXiv:2201.05936}, 2022.

\bibitem{LiuISIT2021}
J.~Liu, W.~Zhang, and H.~V. Poor,
\newblock ``{A Rate-Distortion Framework for Characterizing Semantic
  Information},''
\newblock in {\em {IEEE International Symposium on Information Theory (ISIT)}},
  Grenoble, France, Jul. 2021.

\bibitem{FarshbafanArxiv2022}
M.~K. Farshbafan, W.~Saad, and M.~Debbah,
\newblock ``{Curriculum Learning for Goal-Oriented Semantic Communications with
  a Common Language},''
\newblock {\em arXiv preprint arXiv:2111.08051}, Feb. 2022.

\bibitem{MerluzziEuCNC2022}
M.~Merluzzi, M.~C. Filippou, L.~G. Baltar, and E.~C. Strinati,
\newblock ``{Effective Goal-oriented 6G Communications: the Energy-aware Edge
  Inferencing Case},''
\newblock in {\em IEEE EuCNC and 6G Summit}, Grenoble, France, Jun. 2022.

\bibitem{GarcezArxiv2020}
A.~A. Garcez and L.~C. Lamb,
\newblock ``{Neurosymbolic AI: the 3rd Wave},''
\newblock {\em arXiv preprint arXiv:2012.05876}, 2020.

\bibitem{RaedtIJCAI2020}
L.~D. Raedt, S.~Duman\v ci\'c, Robin Manhaeve, and Giuseppe Marra,
\newblock ``{From Statistical Relational to Neural-Symbolic Artificial
  Intelligence},''
\newblock in {\em Proceedings of the Twenty-Ninth International Joint
  Conference on Artificial Intelligence (IJCAI-20) Survey Track}, Yokohama,
  Japan, 2020.

\bibitem{BengioNeurIPS2021}
E.~Bengio, M.~Jain, M.~Korablyov, D.~Precup, and Y.~Bengio.,
\newblock ``{Flow Network based Generative Models for Non-Iterative Diverse
  Candidate Generation},''
\newblock in {\em Neural Information Processing Systems}, Dec. 2021.

\bibitem{DeleuArxiv2022}
T.~Deleu, A.~Grois, C.~Emezue, M.~Rankawat, S.~Lacoste-Julien, S.~Bauer, and
  Y.~Bengio,
\newblock ``{Bayesian Structure Learning with Generative Flow Networks},''
\newblock {\em arXiv preprint arXiv:2202.13903}, Feb. 2022.

\bibitem{Carnap52}
R.~Carnap and Y.~Bar-Hillel,
\newblock ``{An Outline of a Theory of Semantic Information},''
\newblock {\em Technical Report No. 247}, Oct. 1952.

\bibitem{Badreddine2022}
S.~Badreddine, A.~D.~A. Garcez, L.~Serafini, and M.~Spranger,
\newblock ``{Logic Tensor Networks},''
\newblock {\em Artificial Intelligence}, vol. 303, Oct. 2022.

\bibitem{TOSheaTCN2017}
T.~O'Shea and J.~Hoydis,
\newblock ``{An Introduction to Deep Learning for the Physical Layer },''
\newblock {\em IEEE Transactions on Cognitive Communications and Networking},
  vol. 3, no. 4, pp. 563--575, 2017.

\end{thebibliography}
